\documentclass[11pt]{article}

\usepackage[preprint]{acl}

\usepackage{times}
\usepackage{latexsym}

\usepackage[T1]{fontenc}

\usepackage[utf8]{inputenc}

\usepackage{microtype}

\usepackage{inconsolata}

\usepackage{graphicx}
\usepackage{booktabs} 
\usepackage{multirow} 

\usepackage{amsmath}

\usepackage{amssymb}

\usepackage{arydshln}
\usepackage{enumitem}
\usepackage{ragged2e}

\title{Learning from Emptiness: De-biasing Listwise Rerankers with Content-Agnostic Probability Calibration}

\author{%
   Hang Lv\textsuperscript{1}\footnotemark[1],
   Hongchao Gu\textsuperscript{1}\footnotemark[1],
   Ruiqing Yang\textsuperscript{1},
   Liangyue Li\textsuperscript{2},
   Zulong Chen\textsuperscript{2},\\
   \textbf{
   Defu Lian\textsuperscript{1},
   Hao Wang\textsuperscript{1}\footnotemark[2],
   Enhong Chen\textsuperscript{1}}
   \\
   \textsuperscript{1}University of Science and Technology of China \quad
   \textsuperscript{2}Alibaba Group
}

\begin{document}
\maketitle

\begingroup
\renewcommand{\thefootnote}{\fnsymbol{footnote}}
\footnotetext[1]{Equal contribution. Work partially done during an internship at Alibaba Group.}
\footnotetext[2]{Corresponding author: \texttt{wanghao3@ustc.edu.cn}.}
\endgroup
\setcounter{footnote}{0}

\begin{abstract}

Generative listwise reranking leverages global context for superior retrieval but is plagued by intrinsic position bias, where models exhibit structural sensitivity to input order independent of relevance. Existing mitigations present a dilemma: inference-time aggregation incurs prohibitive latency, while training-based methods often fail to eradicate ingrained priors, particularly in compact models. To resolve this dilemma, we propose CapCal (Content-Agnostic Probability Calibration), a training-free framework that mechanically decouples positional bias from ranking decisions. By estimating the bias distribution via content-free placeholders, CapCal rectifies output logits through an entropy-adaptive contrastive mechanism. Evaluations across 10 benchmarks confirm that CapCal achieves superior performance among training-free methods while preserving single-pass efficiency. Notably, it unlocks the latent potential of lightweight models (e.g., 0.6B), delivering absolute NDCG gains exceeding 10 points and outperforming both permutation-based aggregation and data-augmentation baselines.

\end{abstract}

\section{Introduction}

\begin{figure*}[t]
    \centering
    \includegraphics[width=1.02\textwidth]{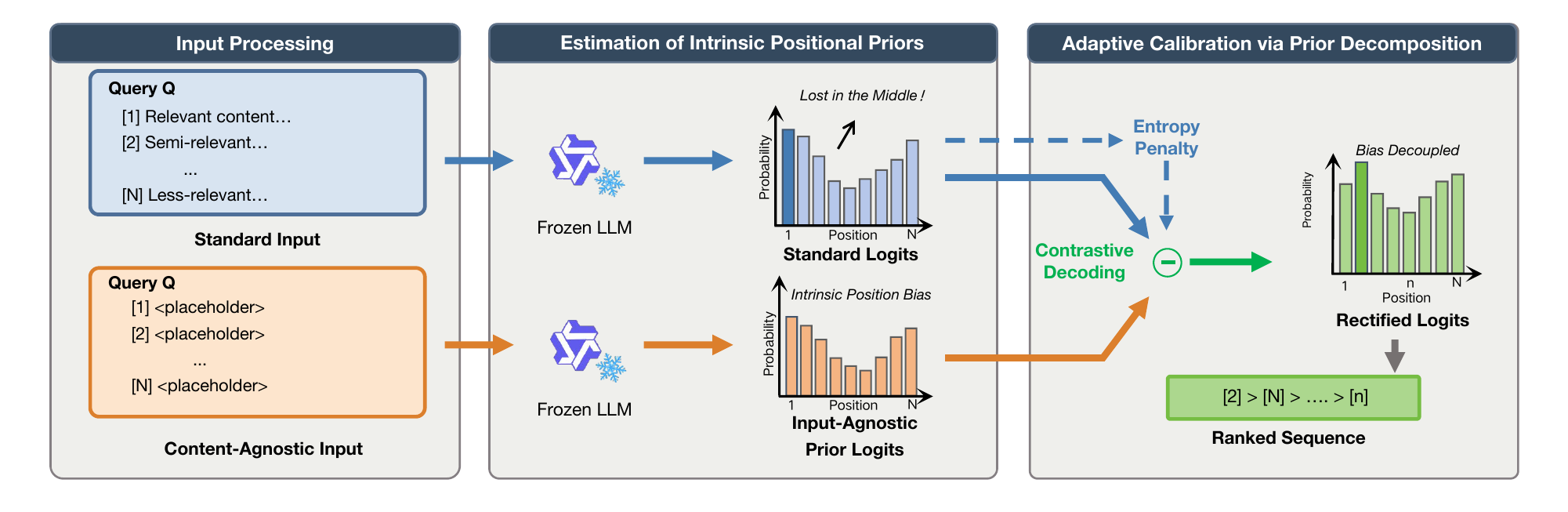}
    
    \caption{\textbf{Overview of the proposed CapCal framework.} The framework decouples position bias by utilizing an empty-passage query to capture the input-agnostic prior. We then apply contrastive decoding to subtract this prior from the standard inference logits, achieving a calibrated ranking.}
    
    \label{fig:main_method}
\end{figure*}



Listwise reranking~\cite{rankgpt, lrl} has established itself as a powerful paradigm in modern information retrieval, utilizing Large Language Models to evaluate candidate documents simultaneously rather than in isolation. By integrating the query and the full document list within a single prompt, this approach enables the model to capture global inter-document dependencies, delivering ranking performance that substantially outperforms traditional pointwise or pairwise methods.


However, this paradigm faces a fundamental theoretical flaw: the violation of \textit{Permutation Invariance}. Theoretically, an ideal ranking function must yield results that are invariant to the physical sequence of candidate documents. In practice, however, LLMs exhibit intrinsic sensitivity to token positions, inevitably giving rise to severe \textbf{position bias}. This often manifests as the "Lost in the Middle" phenomenon~\cite{lost_in_the_middle}, where models systematically favor documents at the beginning or end of the context while overlooking highly relevant ones in the middle, thereby degrading the overall reliability of the retrieval system.

While extensive research has been dedicated to mitigating this bias, existing methodologies remain constrained by an inherent trade-off between efficiency and flexibility. On one hand, inference-time aggregation strategies~\cite{psc,moi} attempt to neutralize bias by averaging predictions across multiple input permutations; however, the computational overhead incurs prohibitive latency, rendering them impractical for real-time scenarios. On the other hand, training-based interventions~\cite{rankvicuna,scalr} seek to enforce order invariance through extensive data augmentation or architectural modifications. Yet, these approaches introduce significant rigidity: they necessitate costly retraining for any new backbone and often exhibit suboptimal cross-domain generalization. Consequently, there remains a critical need for an efficient, training-free solution capable of rectifying position bias without succumbing to the computational burden of aggregation or the inflexibility of retraining.

In this work, we propose CapCal (\textbf{C}ontent-\textbf{A}gnostic \textbf{P}robability \textbf{Cal}ibration), a novel framework that resolves this dilemma by fundamentally reconceptualizing the nature of position bias. Rather than treating bias merely as a symptom to be suppressed, we formulate it as a measurable component to be explicitly decoupled. As illustrated in Figure~\ref{fig:main_method}, our approach is grounded in the key insight that intrinsic priors can be isolated by observing the model's behavior in the absence of content. Specifically, by querying the model with empty passages, we capture the content-agnostic prior---a baseline preference for specific list indices manifested directly in the logits. We then leverage contrastive calibration to analytically subtract this prior from the standard generation probabilities. This strategy effectively rectifies ranking decisions at the distribution level, offering a robust, plug-and-play solution that eliminates position bias without compromising the inference efficiency of the listwise paradigm. We validate CapCal across 10 benchmarks (MS MARCO and BEIR) using Qwen models ranging from 0.6B to 8B. Empirical results demonstrate that our framework consistently enhances ranking performance while maintaining single-pass efficiency. Notably, it unlocks the potential of smaller models, delivering over 10 NDCG points of improvement in high-bias scenarios and proving robust across diverse retrieval contexts.

Our contributions are summarized as follows: 
\begin{itemize}[leftmargin=1em,labelsep=0.4em,itemsep=0.2ex,topsep=0.3ex,parsep=0pt,partopsep=0pt]
\item We introduce CapCal, a plug-and-play framework that mechanically rectifies position bias via content-agnostic probability calibration, eliminating the need for retraining or expensive inference aggregation. 
\item We bridge the capability gap for small language models, demonstrating that explicit bias correction allows 0.6B-scale models to rival larger baselines, thereby enhancing the scalability of generative reranking. 
\item We reveal the limitations of implicit adaptation, showing that inference-time calibration is superior to permutation-based data augmentation in handling the stubborn, intrinsic nature of positional priors. 
\end{itemize}



\section{Related Work}
\label{sec:related_work}

\subsection{Generative listwise reranking}

Generative listwise reranking has superseded pointwise and pairwise paradigms by enabling LLMs to capture global inter-document dependencies. \citet{rankgpt} and \citet{lrl} pioneered this approach by prompting LLMs to directly generate document permutations. Subsequent efforts, such as \citet{rankvicuna} and \citet{pradeep2023rankzephyr}, optimized this via instruction fine-tuning on open-source backbones, while \citet{first} further reduced latency by deriving rankings solely from the first token's logits. Although recent trends explore integrating reasoning steps to enhance precision\cite{reasonrank,rank1,rankk}, such methods incur severe latency costs due to extensive token generation. Consequently, direct generation with efficient, smaller-scale models remains a practical choice for latency-sensitive applications, making robust position-bias mitigation especially important.

\subsection{Position Bias and Mitigation Strategies}

LLMs exhibit intrinsic sensitivity to input order, leading to the "Lost in the Middle" phenomenon \cite{lost_in_the_middle}. Mitigation strategies primarily diverge into inference-time input manipulation and model-level adaptation. \textbf{Inference-time approaches} treat the LLM as a black box. Aggregation strategies, such as Permutation Self-Consistency (PSC) \cite{psc}, LLM-RankFusion \cite{rankfusion}, and Mixture-of-Intervention (MOI) \cite{moi}, neutralize positional noise by averaging or solving for rankings across multiple shuffled inputs. Decomposition strategies like sliding windows \cite{rankgpt} or tournament sorting \cite{listt5} mitigate bias by segmenting lists to limit context length. While effective, these methods incur prohibitive latency due to repeated inference. Conversely, \textbf{model-level approaches} target internal mechanisms. Data augmentation implies training on shuffled samples to learn invariance\cite{rankvicuna,pradeep2023rankzephyr}, while architectural modifications---such as ListT5's \cite{listt5} ranking loss or SCaLR's \cite{scalr} dual-view alignment---enforce it structurally. However, these techniques often suffer from inflexibility and potential generalization issues due to costly retraining. Lastly, calibration methods like ICR \cite{icr} and FitM \cite{fitm} leverage internal attention weights to mathematically decouple positional priors. While sharing the goal of parameter-free bias isolation, these methods primarily focus on normalizing attention maps rather than rectifying generative distribution.



\subsection{Decoding Strategies and Calibration}
Contrastive Decoding (CD) \cite{contrastive,contrastive2} was originally proposed to enhance generation quality by maximizing the divergence between an "expert" and an "amateur" model, effectively suppressing generic or repetitive modes. To eliminate the need for auxiliary models, self-contrastive approaches \cite{dola} derive the negative constraint from the model itself, typically via unconditional prompts\cite{cot_cd} or early-exit layers\cite{distill_cd}. In Retrieval-Augmented Generation, methods like Context-Aware Decoding (CAD) \cite{CAD,CDRAG2,entropy_CD} apply this principle to amplify contextual reliance by subtracting context-agnostic priors. Our work extends this idea to listwise reranking: we treat the empty-context prior as the mathematical manifestation of positional bias and subtract it directly from the ranking distribution, yielding a training-free debiasing mechanism tailored to generative ranking.

\section{Methodology}

We present \textbf{CapCal}, a training-free framework designed to calibrate listwise reranking by explicitly decoupling inherent positional priors from semantic relevance. As illustrated in Figure \ref{fig:main_method}, our approach operates in two stages: quantifying the input-agnostic prior via semantic-free placeholders, and rectifying the ranking distribution through an adaptive, contrastive calibration mechanism.

\subsection{Preliminaries: Listwise Reranking}

Given a query $q$ and a list of $N$ candidate documents $\mathcal{D} = \{d_1, d_2, \dots, d_N\}$, the goal is to generate a permutation $\pi$ of indices that ranks documents by decreasing relevance. Following the standard generative paradigm ~\cite{rankgpt}, the input prompt $x$ is constructed by concatenating the query and the document list: $x = \mathcal{T}(q, d_1, \dots, d_N)$, where $\mathcal{T}$ is a prompt template as shown in \ref{sec:appendix}. At step $k$, the probability of generating the identifier token $t_k$ corresponding to document $d_i$ is computed as: $P(t_k = d_i \mid x, t_{<k}) = \frac{\exp(\mathbf{z}_{k, d_i})}{\sum{v \in \mathcal{V}} \exp(\mathbf{z}{k, v})}$, where $\mathbf{z}_{k} \in \mathbb{R}^{|\mathcal{V}|}$ are the logits over the vocabulary V.


\subsection{Estimation of Intrinsic Positional Priors}


Position bias intrinsically manifests as a non-uniform probability distribution over document indices, persisting independently of semantic content. To isolate this bias, we construct a content-free input $x_{\text{empty}}$. Specifically, we mirror the standard prompt structure but replace the textual content of each candidate document with a semantically vacuous $\varnothing$ (e.g., an empty string), while retaining the original query and document identifiers:
\begin{equation}
    x_{\text{empty}} = \mathcal{T}(q, \varnothing, \dots, \varnothing)
\end{equation}
By feeding $x_{\text{empty}}$ into the frozen LLM, we obtain the prior logits $\mathbf{z}^{\text{prior}}_k$. Since the candidate list is devoid of semantic information, any divergence in the probability mass assigned to different document indices (e.g., a structural preference for "1" over "5") serves as a direct quantification of the model's intrinsic positional bias.

\subsection{Adaptive Calibration via Prior Decomposition}


Standard contrastive decoding methods operate on token-level logits, which is ill-suited for listwise reranking where document identifiers vary in token length (e.g., "9" vs. "10"). Consequently, we formulate our calibration strategy directly within the \textbf{probability space}, focusing on alignment, decomposition, and adaptive correction.

\paragraph{Identifier-Level Probability Alignment.}

First, we estimate the generation probability for each candidate document $d_i$. To normalize across identifiers of varying lengths, we compute the joint probability of the constituent token sequence $(w_1, \dots, w_m)$ appended with a termination symbol (e.g., `]`) to normalize across lengths:
\begin{equation}
    P(d_i \mid x) = \prod_{j=1}^{m} P(w_j \mid x, w_{<j}) \cdot P(\text{]} \mid x, w_{1:m})
    \label{eq:joint_prob}
\end{equation}



\paragraph{Decomposition of Positional Bias.}
We define the calibrated score $S(d_i)$ by decomposing the model's prediction into a semantic component and a bias component. Theoretically, in the complete absence of semantic signals, an ideal unbiased ranker must yield a uniform distribution over the candidate set. Deviations from this uniform baseline constitute excessive prior bias.
The calibrated score is derived by subtracting this bias from the original probability:
\begin{equation}
    S(d_i) = P(d_i \mid x) - \alpha \cdot \left( P(d_i \mid x_{\text{empty}}) - \frac{1}{|\mathcal{C}_k|} \right)
    \label{eq:calibration}
\end{equation}
where $\mathcal{C}_k$ denotes the set of available candidates at step $k$, and $\frac{1}{|\mathcal{C}_k|}$ represents the theoretical uniform assumption. 



\paragraph{Entropy-Based Adaptive Penalty.}
Instead of a static penalty, we introduce a dynamic coefficient $\alpha_k$ to adapt to the model's fluctuating uncertainty. This dynamic adjustment is motivated by the observation that LLMs tend to fall back on positional shortcuts primarily when semantic confidence diminishes. Accordingly, we modulate $\alpha_k$ based on the Shannon entropy of the prediction distribution:
\begin{equation}
    \alpha_k = \beta \cdot \mathcal{H}(P(\cdot \mid x, t_{<k})) = -\beta \sum_{d \in \mathcal{C}_k} p(d) \log p(d)
\end{equation}
where $p(d)$ is the normalized probability of candidate $d$, and $\beta$ is a scaling hyperparameter. This ensures strong calibration when uncertainty is high (flat distribution), while preserving original semantics when the model is confident.

\paragraph{Constrained Decoding.}
Finally, to guarantee the validity of the output ranking, we employ constrained decoding. At each step, the vocabulary is restricted to the valid next tokens corresponding to indices in $\mathcal{C}_k$, ensuring the generated sequence forms a valid permutation of the input documents.
\section{Experiments}

\begin{table*}[t]
\centering
\small
\setlength{\tabcolsep}{3.0pt} 
\begin{tabular}{ll ccccc c ccccc c}
\toprule
 & & \multicolumn{2}{c}{\textbf{MSv1}} & \multicolumn{3}{c}{\textbf{MSv2}} & \textbf{Avg.} & \multicolumn{5}{c}{\textbf{BEIR}} & \textbf{Avg.} \\
\cmidrule(lr){3-4} \cmidrule(lr){5-7} \cmidrule(lr){9-13}
\textbf{Model} & \textbf{Method} & \textbf{DL19} & \textbf{DL20} & \textbf{DL21} & \textbf{DL22} & \textbf{DL23} & \textbf{InD} & \textbf{COVID} & \textbf{Clim.} & \textbf{NF} & \textbf{Argu.} & \textbf{FiQA} & \textbf{OOD} \\
\midrule
\multirow{2}{*}{Qwen3-0.6b} & Base & 0.4916 & 0.3415 & 0.6278 & 0.4858 & 0.5457 & 0.4985 & 0.6074 & 0.1622 & 0.4243 & 0.1097 & 0.1807 & 0.2969 \\
                            & CapCal & \textbf{0.5454} & \textbf{0.4126} & \textbf{0.6666} & \textbf{0.5457} & \textbf{0.5762} & \textbf{0.5493} & \textbf{0.6400} & \textbf{0.2086} & \textbf{0.4504} & \textbf{0.2448} & \textbf{0.2364} & \textbf{0.3560} \\
\midrule
\multirow{2}{*}{Qwen3-8b}   & Base & 0.6807 & 0.5405 & 0.8091 & 0.6129 & 0.6945 & 0.6675 & 0.7525 & 0.2486 & 0.4981 & 0.2436 & 0.2877 & 0.4061 \\
                            & CapCal & \textbf{0.7141} & \textbf{0.5613} & \textbf{0.8228} & \textbf{0.6391} & \textbf{0.6979} & \textbf{0.6870} & \textbf{0.7683} & \textbf{0.2714} & \textbf{0.5094} & \textbf{0.2464} & \textbf{0.2932} & \textbf{0.4177} \\
\midrule
\multirow{2}{*}{Qwen2.5-7b} & Base & \textbf{0.7114} & \textbf{0.5151} & 0.8142 & 0.6293 & 0.6317 & \textbf{0.6603} & 0.7542 & 0.2325 & 0.4858 & 0.2181 & 0.2846 & 0.3950 \\
                            & CapCal & 0.6991 & 0.5020 & \textbf{0.8277} & \textbf{0.6417} & \textbf{0.6411} & 0.6623 & \textbf{0.7665} & \textbf{0.2405} & \textbf{0.5088} & \textbf{0.2553} & \textbf{0.3075} & \textbf{0.4157} \\
\bottomrule
\end{tabular}
\caption{Performance comparison on MS MARCO and BEIR benchmarks. All metrics are reported in NDCG@10.}
\label{tab:main}
\end{table*}

\subsection{Experiment Setup}


\paragraph{Tasks.} We conduct a comprehensive evaluation on 10 benchmarks spanning general web search and specialized domains. For \textit{General Web Search}, we adhere to the standard MS MARCO \cite{bajaj2018msmarcohumangenerated} protocols, covering the v1 corpus with TREC DL19 and DL20 \cite{craswell2020overviewtrec2019deep,craswell2021overviewtrec2020deep} and the v2 corpus with TREC DL21--DL23 \cite{craswell2025overviewtrec2021deep,craswell2025overviewtrec2022deep,craswell2025overviewtrec2023deep}.\footnote{One reviewer pointed out that the evaluation on TREC DL21 may be less reliable; we therefore retain its results for completeness but treat them as reference only.} For \textit{Specialized Domains}, we adopt a diversity-first selection strategy over BEIR, choosing five datasets from distinct semantic fields: Epidemiology (TREC-COVID), Nutritional Science (NFCorpus), Earth Science (Climate-FEVER), Finance (FiQA), and Argumentation (Arguana). This yields a balanced suite of five general-domain and five specialized-domain benchmarks.

\paragraph{Models.} To validate the scalability and robustness of our framework, we employ models across varying sizes and generations. We utilize the latest \textit{Qwen3} series \cite{yang2025qwen3technicalreport}, specifically the 0.6B and 8B variants, to test efficacy on the latest architectural advancements. Additionally, we incorporate \textit{Qwen2.5-7B-Instruct} \cite{qwen2025qwen25technicalreport} to serve as a strong baseline representing widely deployed instruction-tuned models. 

\subsection{Main Results}


Table \ref{tab:main} presents the comparative analysis of our framework across three distinct model scales and ten diverse benchmarks, together with in-domain and out-of-domain averages for quick comparison. The empirical evidence demonstrates that our training-free calibration mechanism consistently enhances generative ranking performance across almost all settings.


\paragraph{Universal Efficacy Across Scales.} Our framework yields robust gains across all tested architectures, ranging from standard 7B/8B models to the highly compact 0.6B variant. While effective universally, the impact is particularly pronounced on Qwen3-0.6B, achieving absolute improvements exceeding 10 NDCG@10 points on MS MARCO V2 and Climate-Fever. This indicates that while bias rectification benefits models of all sizes, it is instrumental in unlocking the latent ranking potential of smaller models, which are intrinsically more susceptible to positional sensitivity.


\paragraph{Cross-Domain Generalization.} The method exhibits remarkable stability across both general web search and specialized domains. By explicitly decoupling intrinsic positional priors, our calibration restores ranking fidelity, effectively handling zero-shot scenarios and complex reasoning tasks such as TREC DL23 and NFCorpus. The sustained improvements across both MS MARCO settings and diverse BEIR tasks further attest to the domain-agnostic nature of our adaptive penalty mechanism.

\paragraph{Comparison with Inference-Time Aggregation.} We further compare CapCal against Permutation Self-Consistency (PSC), a strong inference-time aggregation baseline that performs 10 shuffled reranking passes per query. As detailed in Appendix~\ref{sec:B.4}, CapCal consistently matches or exceeds PSC on Qwen3-0.6B while requiring only one additional forward pass, demonstrating a substantially better effectiveness-efficiency trade-off.


\subsection{Discussion and Analysis} We employ Qwen3-0.6B to evaluate our framework's resilience and compare it with both inference-time aggregation and training-based debiasing strategies. 


\paragraph{Robustness Analysis.} We challenge our calibration mechanism under three perturbation settings to verify its stability: 
\vspace{-2mm}
\begin{itemize}

\item \textbf{Input Distribution Shift:} We altered the input distribution by employing a stronger first-stage retriever (\textit{bge-reranker-v2-m3} \cite{chen2024bge}) and applying \textit{Random Shuffling} to the candidate list. Our method yields consistent gains regardless of initial retrieval quality or randomized ordering, demonstrating its robustness to input variations and its orthogonality to document sequence. (Appendix ~\ref{sec:B.1})

\vspace{-2mm}
\item \textbf{Placeholder Sensitivity:} We varied both the semantic content and the length of the placeholders used to capture priors. The calibration remains effective across these variations, confirming that the captured bias is structural---stemming from the position indices themselves---rather than being triggered by specific semantic tokens in the dummy input. (Appendix ~\ref{sec:B.2})
\vspace{-2mm}
\item \textbf{Identifier Semantics:} We shifted document identifiers from standard numerals (e.g., "1") to alphabets (e.g., "A"). The framework successfully generalizes to these alternative formats, demonstrating adaptability to diverse prompting strategies. (Appendix ~\ref{sec:B.3})
\end{itemize}


\paragraph{Persistence of Bias in Data Augmentation.} Seminal works such as RankZephyr \cite{pradeep2023rankzephyr} rely on permutation-based data augmentation during training to suppress position bias. We replicated this strategy to fine-tune Qwen3-0.6B. However, empirical results reveal that severe position bias persists even after such targeted training. This suggests that for compact models with limited capacity, positional priors are ingrained and resistant to implicit correction through data diversity alone. In contrast, our inference-time calibration offers a rigorous solution that explicitly rectifies these residual biases where standard data augmentation falls short. (Appendix ~\ref{sec:B.5})

\section{Conclusion}



To address position bias without the latency of aggregation or costs of retraining, we introduce \textbf{CapCal}. This training-free framework calibrates logits using placeholders to isolate bias. Across 10 benchmarks, it achieves better results with single-pass efficiency, notably boosting 0.6B models by more than 10 NDCG points and outperforming both permutation-based aggregation and data-augmentation baselines.

\section*{Acknowledgments}
We sincerely thank the ACL Rolling Review reviewers for their valuable suggestions.

\section*{Limitations}
We acknowledge that CapCal involves two primary limitations. First, the framework introduces a marginal increase in computational overhead because it requires an additional forward pass to isolate the input-agnostic prior using content-free placeholders. While this remains significantly more efficient than permutation-based aggregation strategies, it is inherently slower than standard single-pass generative reranking. Second, because our calibration is performed directly in the probability space, the method requires access to identifier-level probabilities or logits. Consequently, CapCal is not applicable to completely black-box APIs that expose only raw text completions, although it remains compatible with deployment settings that provide log-probabilities or output distributions.

\bibliography{custom}

@inproceedings{rankgpt,
    title = "Is {C}hat{GPT} Good at Search? Investigating Large Language Models as Re-Ranking Agents",
    author = "Sun, Weiwei  and
      Yan, Lingyong  and
      Ma, Xinyu  and
      Wang, Shuaiqiang  and
      Ren, Pengjie  and
      Chen, Zhumin  and
      Yin, Dawei  and
      Ren, Zhaochun",
    editor = "Bouamor, Houda  and
      Pino, Juan  and
      Bali, Kalika",
    booktitle = "Proceedings of the 2023 Conference on Empirical Methods in Natural Language Processing",
    month = dec,
    year = "2023",
    address = "Singapore",
    publisher = "Association for Computational Linguistics",
    url = "https://aclanthology.org/2023.emnlp-main.923/",
    doi = "10.18653/v1/2023.emnlp-main.923",
    pages = "14918--14937"
}

@ARTICLE{rankvicuna,
  title   = {{RankVicuna}: Zero-Shot Listwise Document Reranking with Open-Source Large Language Models},
  author  = {Ronak Pradeep and Sahel Sharifymoghaddam and Jimmy Lin},
  year    = {2023},
  journal = {arXiv:2309.15088}
}

@ARTICLE{pradeep2023rankzephyr,
  title   = {{RankZephyr}: Effective and Robust Zero-Shot Listwise Reranking is a Breeze!},
  author  = {Ronak Pradeep and Sahel Sharifymoghaddam and Jimmy Lin},
  year    = {2023},
  journal = {arXiv:2312.02724}
}

@inproceedings{first,
    title = "{FIRST}: Faster Improved Listwise Reranking with Single Token Decoding",
    author = "Gangi Reddy, Revanth  and
      Doo, JaeHyeok  and
      Xu, Yifei  and
      Sultan, Md Arafat  and
      Swain, Deevya  and
      Sil, Avirup  and
      Ji, Heng",
    editor = "Al-Onaizan, Yaser  and
      Bansal, Mohit  and
      Chen, Yun-Nung",
    booktitle = "Proceedings of the 2024 Conference on Empirical Methods in Natural Language Processing",
    month = nov,
    year = "2024",
    address = "Miami, Florida, USA",
    publisher = "Association for Computational Linguistics",
    url = "https://aclanthology.org/2024.emnlp-main.491/",
    doi = "10.18653/v1/2024.emnlp-main.491",
    pages = "8642--8652"
}

@article{lost_in_the_middle,
    title = "Lost in the Middle: How Language Models Use Long Contexts",
    author = "Liu, Nelson F.  and
      Lin, Kevin  and
      Hewitt, John  and
      Paranjape, Ashwin  and
      Bevilacqua, Michele  and
      Petroni, Fabio  and
      Liang, Percy",
    journal = "Transactions of the Association for Computational Linguistics",
    volume = "12",
    year = "2024",
    address = "Cambridge, MA",
    publisher = "MIT Press",
    url = "https://aclanthology.org/2024.tacl-1.9/",
    doi = "10.1162/tacl_a_00638",
    pages = "157--173"
}

@inproceedings{psc,
    title = "Found in the Middle: Permutation Self-Consistency Improves Listwise Ranking in Large Language Models",
    author = "Tang, Raphael  and
      Zhang, Crystina  and
      Ma, Xueguang  and
      Lin, Jimmy  and
      Ture, Ferhan",
    editor = "Duh, Kevin  and
      Gomez, Helena  and
      Bethard, Steven",
    booktitle = "Proceedings of the 2024 Conference of the North American Chapter of the Association for Computational Linguistics: Human Language Technologies (Volume 1: Long Papers)",
    month = jun,
    year = "2024",
    address = "Mexico City, Mexico",
    publisher = "Association for Computational Linguistics",
    url = "https://aclanthology.org/2024.naacl-long.129/",
    doi = "10.18653/v1/2024.naacl-long.129",
    pages = "2327--2340"
}

@inproceedings{moi,
    title = "Inference Scaling for Bridging Retrieval and Augmented Generation",
    author = "Lee, Youngwon  and
      Hwang, Seung-won  and
      Campos, Daniel F  and
      Grali{\'n}ski, Filip  and
      Yao, Zhewei  and
      He, Yuxiong",
    editor = "Chiruzzo, Luis  and
      Ritter, Alan  and
      Wang, Lu",
    booktitle = "Findings of the Association for Computational Linguistics: NAACL 2025",
    month = apr,
    year = "2025",
    address = "Albuquerque, New Mexico",
    publisher = "Association for Computational Linguistics",
    url = "https://aclanthology.org/2025.findings-naacl.409/",
    doi = "10.18653/v1/2025.findings-naacl.409",
    pages = "7324--7339",
    ISBN = "979-8-89176-195-7"
}

@inproceedings{listt5,
    title = "{L}ist{T}5: Listwise Reranking with Fusion-in-Decoder Improves Zero-shot Retrieval",
    author = "Yoon, Soyoung  and
      Choi, Eunbi  and
      Kim, Jiyeon  and
      Yun, Hyeongu  and
      Kim, Yireun  and
      Hwang, Seung-won",
    editor = "Ku, Lun-Wei  and
      Martins, Andre  and
      Srikumar, Vivek",
    booktitle = "Proceedings of the 62nd Annual Meeting of the Association for Computational Linguistics (Volume 1: Long Papers)",
    month = aug,
    year = "2024",
    address = "Bangkok, Thailand",
    publisher = "Association for Computational Linguistics",
    url = "https://aclanthology.org/2024.acl-long.125/",
    doi = "10.18653/v1/2024.acl-long.125",
    pages = "2287--2308"
}

@misc{LRL,
      title={Zero-Shot Listwise Document Reranking with a Large Language Model}, 
      author={Xueguang Ma and Xinyu Zhang and Ronak Pradeep and Jimmy Lin},
      year={2023},
      eprint={2305.02156},
      archivePrefix={arXiv},
      primaryClass={cs.IR},
      url={https://arxiv.org/abs/2305.02156}, 
}

@misc{ICR,
      title={Attention in Large Language Models Yields Efficient Zero-Shot Re-Rankers}, 
      author={Shijie Chen and Bernal Jiménez Gutiérrez and Yu Su},
      year={2025},
      eprint={2410.02642},
      archivePrefix={arXiv},
      primaryClass={cs.CL},
      url={https://arxiv.org/abs/2410.02642}, 
}

@misc{rankfusion,
      title={LLM-RankFusion: Mitigating Intrinsic Inconsistency in LLM-based Ranking}, 
      author={Yifan Zeng and Ojas Tendolkar and Raymond Baartmans and Qingyun Wu and Lizhong Chen and Huazheng Wang},
      year={2024},
      eprint={2406.00231},
      archivePrefix={arXiv},
      primaryClass={cs.IR},
      url={https://arxiv.org/abs/2406.00231}, 
}

@misc{scalr,
      title={Self-Calibrated Listwise Reranking with Large Language Models}, 
      author={Ruiyang Ren and Yuhao Wang and Kun Zhou and Wayne Xin Zhao and Wenjie Wang and Jing Liu and Ji-Rong Wen and Tat-Seng Chua},
      year={2024},
      eprint={2411.04602},
      archivePrefix={arXiv},
      primaryClass={cs.IR},
      url={https://arxiv.org/abs/2411.04602}, 
}

@misc{fitm,
      title={Found in the Middle: Calibrating Positional Attention Bias Improves Long Context Utilization}, 
      author={Cheng-Yu Hsieh and Yung-Sung Chuang and Chun-Liang Li and Zifeng Wang and Long T. Le and Abhishek Kumar and James Glass and Alexander Ratner and Chen-Yu Lee and Ranjay Krishna and Tomas Pfister},
      year={2024},
      eprint={2406.16008},
      archivePrefix={arXiv},
      primaryClass={cs.CL},
      url={https://arxiv.org/abs/2406.16008}, 
}

@misc{reasonrank,
      title={ReasonRank: Empowering Passage Ranking with Strong Reasoning Ability}, 
      author={Wenhan Liu and Xinyu Ma and Weiwei Sun and Yutao Zhu and Yuchen Li and Dawei Yin and Zhicheng Dou},
      year={2025},
      eprint={2508.07050},
      archivePrefix={arXiv},
      primaryClass={cs.IR},
      url={https://arxiv.org/abs/2508.07050}, 
}

@misc{rank1,
      title={Rank1: Test-Time Compute for Reranking in Information Retrieval}, 
      author={Orion Weller and Kathryn Ricci and Eugene Yang and Andrew Yates and Dawn Lawrie and Benjamin Van Durme},
      year={2025},
      eprint={2502.18418},
      archivePrefix={arXiv},
      primaryClass={cs.IR},
      url={https://arxiv.org/abs/2502.18418}, 
}

@misc{rankk,
      title={Rank-K: Test-Time Reasoning for Listwise Reranking}, 
      author={Eugene Yang and Andrew Yates and Kathryn Ricci and Orion Weller and Vivek Chari and Benjamin Van Durme and Dawn Lawrie},
      year={2025},
      eprint={2505.14432},
      archivePrefix={arXiv},
      primaryClass={cs.IR},
      url={https://arxiv.org/abs/2505.14432}, 
}

@inproceedings{contrastive,
    title = "Contrastive Decoding: Open-ended Text Generation as Optimization",
    author = "Li, Xiang Lisa  and
      Holtzman, Ari  and
      Fried, Daniel  and
      Liang, Percy  and
      Eisner, Jason  and
      Hashimoto, Tatsunori  and
      Zettlemoyer, Luke  and
      Lewis, Mike",
    editor = "Rogers, Anna  and
      Boyd-Graber, Jordan  and
      Okazaki, Naoaki",
    booktitle = "Proceedings of the 61st Annual Meeting of the Association for Computational Linguistics (Volume 1: Long Papers)",
    month = jul,
    year = "2023",
    address = "Toronto, Canada",
    publisher = "Association for Computational Linguistics",
    url = "https://aclanthology.org/2023.acl-long.687/",
    doi = "10.18653/v1/2023.acl-long.687",
    pages = "12286--12312"
}

@misc{contrastive2,
      title={Contrastive Decoding Improves Reasoning in Large Language Models}, 
      author={Sean O'Brien and Mike Lewis},
      year={2023},
      eprint={2309.09117},
      archivePrefix={arXiv},
      primaryClass={cs.CL},
      url={https://arxiv.org/abs/2309.09117}, 
}

@misc{dola,
      title={DoLa: Decoding by Contrasting Layers Improves Factuality in Large Language Models}, 
      author={Yung-Sung Chuang and Yujia Xie and Hongyin Luo and Yoon Kim and James Glass and Pengcheng He},
      year={2024},
      eprint={2309.03883},
      archivePrefix={arXiv},
      primaryClass={cs.CL},
      url={https://arxiv.org/abs/2309.03883}, 
}

@misc{distill_cd,
      title={Distillation Contrastive Decoding: Improving LLMs Reasoning with Contrastive Decoding and Distillation}, 
      author={Phuc Phan and Hieu Tran and Long Phan},
      year={2024},
      eprint={2402.14874},
      archivePrefix={arXiv},
      primaryClass={cs.CL},
      url={https://arxiv.org/abs/2402.14874}, 
}

@inproceedings{cot_cd,
    title = "Enhancing Input-Label Mapping in In-Context Learning with Contrastive Decoding",
    author = "Peng, Keqin  and
      Ding, Liang  and
      Ouyang, Yuanxin  and
      Fang, Meng  and
      Yuan, Yancheng  and
      Tao, Dacheng",
    editor = "Che, Wanxiang  and
      Nabende, Joyce  and
      Shutova, Ekaterina  and
      Pilehvar, Mohammad Taher",
    booktitle = "Proceedings of the 63rd Annual Meeting of the Association for Computational Linguistics (Volume 2: Short Papers)",
    month = jul,
    year = "2025",
    address = "Vienna, Austria",
    publisher = "Association for Computational Linguistics",
    url = "https://aclanthology.org/2025.acl-short.77/",
    doi = "10.18653/v1/2025.acl-short.77",
    pages = "997--1004",
    ISBN = "979-8-89176-252-7"
}

@inproceedings{CAD,
    title = "Trusting Your Evidence: Hallucinate Less with Context-aware Decoding",
    author = "Shi, Weijia  and
      Han, Xiaochuang  and
      Lewis, Mike  and
      Tsvetkov, Yulia  and
      Zettlemoyer, Luke  and
      Yih, Wen-tau",
    editor = "Duh, Kevin  and
      Gomez, Helena  and
      Bethard, Steven",
    booktitle = "Proceedings of the 2024 Conference of the North American Chapter of the Association for Computational Linguistics: Human Language Technologies (Volume 2: Short Papers)",
    month = jun,
    year = "2024",
    address = "Mexico City, Mexico",
    publisher = "Association for Computational Linguistics",
    url = "https://aclanthology.org/2024.naacl-short.69/",
    doi = "10.18653/v1/2024.naacl-short.69",
    pages = "783--791"
}

@inproceedings{CDRAG2,
    title = "Enhancing Contextual Understanding in Large Language Models through Contrastive Decoding",
    author = "Zhao, Zheng  and
      Monti, Emilio  and
      Lehmann, Jens  and
      Assem, Haytham",
    editor = "Duh, Kevin  and
      Gomez, Helena  and
      Bethard, Steven",
    booktitle = "Proceedings of the 2024 Conference of the North American Chapter of the Association for Computational Linguistics: Human Language Technologies (Volume 1: Long Papers)",
    month = jun,
    year = "2024",
    address = "Mexico City, Mexico",
    publisher = "Association for Computational Linguistics",
    url = "https://aclanthology.org/2024.naacl-long.237/",
    doi = "10.18653/v1/2024.naacl-long.237",
    pages = "4225--4237"
}

@inproceedings{entropy_CD,
    title = "Entropy-Based Decoding for Retrieval-Augmented Large Language Models",
    author = "Qiu, Zexuan  and
      Ou, Zijing  and
      Wu, Bin  and
      Li, Jingjing  and
      Liu, Aiwei  and
      King, Irwin",
    editor = "Chiruzzo, Luis  and
      Ritter, Alan  and
      Wang, Lu",
    booktitle = "Proceedings of the 2025 Conference of the Nations of the Americas Chapter of the Association for Computational Linguistics: Human Language Technologies (Volume 1: Long Papers)",
    month = apr,
    year = "2025",
    address = "Albuquerque, New Mexico",
    publisher = "Association for Computational Linguistics",
    url = "https://aclanthology.org/2025.naacl-long.236/",
    doi = "10.18653/v1/2025.naacl-long.236",
    pages = "4616--4627",
    ISBN = "979-8-89176-189-6"
}

@misc{craswell2020overviewtrec2019deep,
      title={Overview of the TREC 2019 deep learning track}, 
      author={Nick Craswell and Bhaskar Mitra and Emine Yilmaz and Daniel Campos and Ellen M. Voorhees},
      year={2020},
      eprint={2003.07820},
      archivePrefix={arXiv},
      primaryClass={cs.IR},
      url={https://arxiv.org/abs/2003.07820}, 
}

@misc{craswell2021overviewtrec2020deep,
      title={Overview of the TREC 2020 deep learning track}, 
      author={Nick Craswell and Bhaskar Mitra and Emine Yilmaz and Daniel Campos},
      year={2021},
      eprint={2102.07662},
      archivePrefix={arXiv},
      primaryClass={cs.IR},
      url={https://arxiv.org/abs/2102.07662}, 
}

@misc{craswell2025overviewtrec2021deep,
      title={Overview of the TREC 2021 deep learning track}, 
      author={Nick Craswell and Bhaskar Mitra and Emine Yilmaz and Daniel Campos and Jimmy Lin},
      year={2025},
      eprint={2507.08191},
      archivePrefix={arXiv},
      primaryClass={cs.IR},
      url={https://arxiv.org/abs/2507.08191}, 
}

@misc{craswell2025overviewtrec2022deep,
      title={Overview of the TREC 2022 deep learning track}, 
      author={Nick Craswell and Bhaskar Mitra and Emine Yilmaz and Daniel Campos and Jimmy Lin and Ellen M. Voorhees and Ian Soboroff},
      year={2025},
      eprint={2507.10865},
      archivePrefix={arXiv},
      primaryClass={cs.IR},
      url={https://arxiv.org/abs/2507.10865}, 
}

@misc{craswell2025overviewtrec2023deep,
      title={Overview of the TREC 2023 deep learning track}, 
      author={Nick Craswell and Bhaskar Mitra and Emine Yilmaz and Hossein A. Rahmani and Daniel Campos and Jimmy Lin and Ellen M. Voorhees and Ian Soboroff},
      year={2025},
      eprint={2507.08890},
      archivePrefix={arXiv},
      primaryClass={cs.IR},
      url={https://arxiv.org/abs/2507.08890}, 
}

@misc{bajaj2018msmarcohumangenerated,
      title={MS MARCO: A Human Generated MAchine Reading COmprehension Dataset}, 
      author={Payal Bajaj and Daniel Campos and Nick Craswell and Li Deng and Jianfeng Gao and Xiaodong Liu and Rangan Majumder and Andrew McNamara and Bhaskar Mitra and Tri Nguyen and Mir Rosenberg and Xia Song and Alina Stoica and Saurabh Tiwary and Tong Wang},
      year={2018},
      eprint={1611.09268},
      archivePrefix={arXiv},
      primaryClass={cs.CL},
      url={https://arxiv.org/abs/1611.09268}, 
}

@misc{yang2025qwen3technicalreport,
      title={Qwen3 Technical Report}, 
      author={An Yang and Anfeng Li and Baosong Yang and Beichen Zhang and Binyuan Hui and Bo Zheng and Bowen Yu and Chang Gao and Chengen Huang and Chenxu Lv and Chujie Zheng and Dayiheng Liu and Fan Zhou and Fei Huang and Feng Hu and Hao Ge and Haoran Wei and Huan Lin and Jialong Tang and Jian Yang and Jianhong Tu and Jianwei Zhang and Jianxin Yang and Jiaxi Yang and Jing Zhou and Jingren Zhou and Junyang Lin and Kai Dang and Keqin Bao and Kexin Yang and Le Yu and Lianghao Deng and Mei Li and Mingfeng Xue and Mingze Li and Pei Zhang and Peng Wang and Qin Zhu and Rui Men and Ruize Gao and Shixuan Liu and Shuang Luo and Tianhao Li and Tianyi Tang and Wenbiao Yin and Xingzhang Ren and Xinyu Wang and Xinyu Zhang and Xuancheng Ren and Yang Fan and Yang Su and Yichang Zhang and Yinger Zhang and Yu Wan and Yuqiong Liu and Zekun Wang and Zeyu Cui and Zhenru Zhang and Zhipeng Zhou and Zihan Qiu},
      year={2025},
      eprint={2505.09388},
      archivePrefix={arXiv},
      primaryClass={cs.CL},
      url={https://arxiv.org/abs/2505.09388}, 
}

@misc{qwen2025qwen25technicalreport,
      title={Qwen2.5 Technical Report}, 
      author={Qwen and : and An Yang and Baosong Yang and Beichen Zhang and Binyuan Hui and Bo Zheng and Bowen Yu and Chengyuan Li and Dayiheng Liu and Fei Huang and Haoran Wei and Huan Lin and Jian Yang and Jianhong Tu and Jianwei Zhang and Jianxin Yang and Jiaxi Yang and Jingren Zhou and Junyang Lin and Kai Dang and Keming Lu and Keqin Bao and Kexin Yang and Le Yu and Mei Li and Mingfeng Xue and Pei Zhang and Qin Zhu and Rui Men and Runji Lin and Tianhao Li and Tianyi Tang and Tingyu Xia and Xingzhang Ren and Xuancheng Ren and Yang Fan and Yang Su and Yichang Zhang and Yu Wan and Yuqiong Liu and Zeyu Cui and Zhenru Zhang and Zihan Qiu},
      year={2025},
      eprint={2412.15115},
      archivePrefix={arXiv},
      primaryClass={cs.CL},
      url={https://arxiv.org/abs/2412.15115}, 
}

@misc{chen2024bge,
      title={BGE M3-Embedding: Multi-Lingual, Multi-Functionality, Multi-Granularity Text Embeddings Through Self-Knowledge Distillation}, 
      author={Jianlv Chen and Shitao Xiao and Peitian Zhang and Kun Luo and Defu Lian and Zheng Liu},
      year={2024},
      eprint={2402.03216},
      archivePrefix={arXiv},
      primaryClass={cs.CL}
}

@misc{tunstall2023zephyrdirectdistillationlm,
      title={Zephyr: Direct Distillation of LM Alignment}, 
      author={Lewis Tunstall and Edward Beeching and Nathan Lambert and Nazneen Rajani and Kashif Rasul and Younes Belkada and Shengyi Huang and Leandro von Werra and Clémentine Fourrier and Nathan Habib and Nathan Sarrazin and Omar Sanseviero and Alexander M. Rush and Thomas Wolf},
      year={2023},
      eprint={2310.16944},
      archivePrefix={arXiv},
      primaryClass={cs.LG},
      url={https://arxiv.org/abs/2310.16944}, 
}

@inproceedings{liang-etal-2025-adaptive,
    title = "Adaptive Schema-aware Event Extraction with Retrieval-Augmented Generation",
    author = {Sheng Liang and Hang Lv and Zhihao Wen and Yaxiong Wu and Yongyue Zhang and Hao Wang and Yong Liu},
    booktitle = "Findings of the Association for Computational Linguistics: EMNLP 2025",
    year = "2025",
    url = "https://arxiv.org/abs/2505.08690",
}

@misc{wu-etal-2025-from,
      title={From Human Memory to AI Memory: A Survey on Memory Mechanisms in the Era of LLMs}, 
      author={Yaxiong Wu and Sheng Liang and Chen Zhang and Yichao Wang and Yongyue Zhang and Huifeng Guo and Ruiming Tang and Yong Liu},
      year={2025},
      eprint={2504.15965},
      archivePrefix={arXiv},
      primaryClass={cs.IR},
      url={https://arxiv.org/abs/2504.15965}, 
}

@inproceedings{nie-etal-2023-cross,
    title = "Cross-Lingual Retrieval Augmented Prompt for Low-Resource Languages",
    author = {Nie, Ercong  and  Liang, Sheng  and  Schmid, Helmut  and  Sch{\"u}tze, Hinrich},
    booktitle = "Findings of the Association for Computational Linguistics: ACL 2023",
    year = "2023",
    url = "https://aclanthology.org/2023.findings-acl.528/",
}

@inproceedings{liang-etal-2020-monolingual,
    title = "Monolingual and Multilingual Reduction of Gender Bias in Contextualized Representations",
    author = {Liang, Sheng  and   Dufter, Philipp  and Sch{\"u}tze, Hinrich},
    booktitle = "Proceedings of the 28th International Conference on Computational Linguistics",
    year = "2020",
    url = "https://aclanthology.org/2020.coling-main.446/",
}

@inproceedings{li-etal-2023-from,
    author = {Li, Xiaoqian  and Nie, Ercong  and  Liang, Sheng},
    booktitle = {NeurIPS 2023 Workshop on Instruction Tuning and Instruction Following}, 
    title = {From Classification to Generation: Insights into Crosslingual Retrieval Augmented ICL},
    year = "2023",
    url = "https://arxiv.org/abs/2311.06595"
}

@misc{zhang2025killingbirdsstoneunifying,
      title={Killing Two Birds with One Stone: Unifying Retrieval and Ranking with a Single Generative Recommendation Model}, 
      author={Luankang Zhang and Kenan Song and Yi Quan Lee and Wei Guo and Hao Wang and Yawen Li and Huifeng Guo and Yong Liu and Defu Lian and Enhong Chen},
      year={2025},
      eprint={2504.16454},
      archivePrefix={arXiv},
      primaryClass={cs.IR},
      url={https://arxiv.org/abs/2504.16454}, 
}

@misc{lv2026costeercollaborativedecodingtimepersonalization,
      title={CoSteer: Collaborative Decoding-Time Personalization via Local Delta Steering}, 
      author={Hang Lv and Sheng Liang and Hao Wang and Hongchao Gu and Yaxiong Wu and Wei Guo and Defu Lian and Yong Liu and Enhong Chen},
      year={2026},
      eprint={2507.04756},
      archivePrefix={arXiv},
      primaryClass={cs.CL},
      url={https://arxiv.org/abs/2507.04756}, 
}

@misc{lv2026specsteersynergizinglocalcontext,
      title={SpecSteer: Synergizing Local Context and Global Reasoning for Efficient Personalized Generation}, 
      author={Hang Lv and Sheng Liang and Hao Wang and Yongyue Zhang and Hongchao Gu and Wei Guo and Defu Lian and Yong Liu and Enhong Chen},
      year={2026},
      eprint={2603.16219},
      archivePrefix={arXiv},
      primaryClass={cs.CL},
      url={https://arxiv.org/abs/2603.16219}, 
}

@misc{zhang2026paradigmusercentricagentplatformcentric,
      title={The Next Paradigm Is User-Centric Agent, Not Platform-Centric Service}, 
      author={Luankang Zhang and Hang Lv and Qiushi Pan and Kefen Wang and Yonghao Huang and Xinrui Miao and Yin Xu and Wei Guo and Yong Liu and Hao Wang and Enhong Chen},
      year={2026},
      eprint={2602.15682},
      archivePrefix={arXiv},
      primaryClass={cs.IR},
      url={https://arxiv.org/abs/2602.15682}, 
}

@misc{zhang2026thinkinghurtsdiagnosingrectifying,
      title={Why Thinking Hurts? Diagnosing and Rectifying the Reasoning Shift in Foundation Recommender Models}, 
      author={Luankang Zhang and Yonghao Huang and Hang Lv and Mingjia Yin and Liangyue Li and Zulong Chen and Hao Wang and Enhong Chen},
      year={2026},
      eprint={2602.16587},
      archivePrefix={arXiv},
      primaryClass={cs.IR},
      url={https://arxiv.org/abs/2602.16587}, 
}

@misc{zhang2025ragigbenchinnovativeevaluationragbased,
      title={RAG-IGBench: Innovative Evaluation for RAG-based Interleaved Generation in Open-domain Question Answering}, 
      author={Rongyang Zhang and Yuqing Huang and Chengqiang Lu and Qimeng Wang and Yan Gao and Yi Wu and Yao Hu and Yin Xu and Wei Wang and Hao Wang and Enhong Chen},
      year={2025},
      eprint={2512.05119},
      archivePrefix={arXiv},
      primaryClass={cs.IR},
      url={https://arxiv.org/abs/2512.05119}, 
}

@misc{ye2026fuxilinearunleashingpowerlinear,
      title={FuXi-Linear: Unleashing the Power of Linear Attention in Long-term Time-aware Sequential Recommendation}, 
      author={Yufei Ye and Wei Guo and Hao Wang and Luankang Zhang and Heng Chang and Hong Zhu and Yuyang Ye and Yong Liu and Defu Lian and Enhong Chen},
      year={2026},
      eprint={2602.23671},
      archivePrefix={arXiv},
      primaryClass={cs.IR},
      url={https://arxiv.org/abs/2602.23671}, 
}

@misc{gu2025rapidefficientretrievalaugmentedlong,
      title={RAPID: Efficient Retrieval-Augmented Long Text Generation with Writing Planning and Information Discovery}, 
      author={Hongchao Gu and Dexun Li and Kuicai Dong and Hao Zhang and Hang Lv and Hao Wang and Defu Lian and Yong Liu and Enhong Chen},
      year={2025},
      eprint={2503.00751},
      archivePrefix={arXiv},
      primaryClass={cs.CL},
      url={https://arxiv.org/abs/2503.00751}, 
}

@misc{yin2025featureinteractionfeaturegeneration,
      title={From Feature Interaction to Feature Generation: A Generative Paradigm of CTR Prediction Models}, 
      author={Mingjia Yin and Junwei Pan and Hao Wang and Ximei Wang and Shangyu Zhang and Jie Jiang and Defu Lian and Enhong Chen},
      year={2025},
      eprint={2512.14041},
      archivePrefix={arXiv},
      primaryClass={cs.IR},
      url={https://arxiv.org/abs/2512.14041}, 
}

@misc{shen2024exploringuserretrievalintegration,
      title={Exploring User Retrieval Integration towards Large Language Models for Cross-Domain Sequential Recommendation}, 
      author={Tingjia Shen and Hao Wang and Jiaqing Zhang and Sirui Zhao and Liangyue Li and Zulong Chen and Defu Lian and Enhong Chen},
      year={2024},
      eprint={2406.03085},
      archivePrefix={arXiv},
      primaryClass={cs.LG},
      url={https://arxiv.org/abs/2406.03085}, 
}

@inproceedings{Wang_2025, series={SIGIR ’25},
   title={DLF: Enhancing Explicit-Implicit Interaction via Dynamic Low-Order-Aware Fusion for CTR Prediction},
   url={http://dx.doi.org/10.1145/3726302.3729956},
   DOI={10.1145/3726302.3729956},
   booktitle={Proceedings of the 48th International ACM SIGIR Conference on Research and Development in Information Retrieval},
   publisher={ACM},
   author={Wang, Kefan and Wang, Hao and Guo, Wei and Liu, Yong and Lin, Jianghao and Lian, Defu and Chen, Enhong},
   year={2025},
   month=jul, pages={2213–2223},
   collection={SIGIR ’25} }

@misc{zhi2026spardselfpacedcurriculumrl,
      title={SPARD: Self-Paced Curriculum for RL Alignment via Integrating Reward Dynamics and Data Utility}, 
      author={Xuyang Zhi and Peilun zhou and Chengqiang Lu and Hang Lv and Yiwei Liang and Rongyang Zhang and Yan Gao and YI WU and Yao Hu and Hongchao Gu and Defu Lian and Hao Wang and Enhong Chen},
      year={2026},
      eprint={2604.07837},
      archivePrefix={arXiv},
      primaryClass={cs.AI},
      url={https://arxiv.org/abs/2604.07837}, 
}

@misc{huang2025selfaugmitigatingcatastrophicforgetting,
      title={SelfAug: Mitigating Catastrophic Forgetting in Retrieval-Augmented Generation via Distribution Self-Alignment}, 
      author={Yuqing Huang and Rongyang Zhang and Qimeng Wang and Chengqiang Lu and Yan Gao and Yi Wu and Yao Hu and Xuyang Zhi and Guiquan Liu and Xin Li and Hao Wang and Enhong Chen},
      year={2025},
      eprint={2509.03934},
      archivePrefix={arXiv},
      primaryClass={cs.CL},
      url={https://arxiv.org/abs/2509.03934}, 
}

@misc{cheng2026efficientpersonalizedrerankingsemiautoregressive,
      title={Efficient Personalized Reranking with Semi-Autoregressive Generation and Online Knowledge Distillation}, 
      author={Kai Cheng and Hao Wang and Wei Guo and Weiwen Liu and Yong Liu and Yawen Li and Enhong Chen},
      year={2026},
      eprint={2603.07107},
      archivePrefix={arXiv},
      primaryClass={cs.IR},
      url={https://arxiv.org/abs/2603.07107}, 
}

@misc{wang2025enhancingctrpredictiondecorrelated,
      title={Enhancing CTR Prediction with De-correlated Expert Networks}, 
      author={Jiancheng Wang and Mingjia Yin and Hao Wang and Enhong Chen},
      year={2025},
      eprint={2505.17925},
      archivePrefix={arXiv},
      primaryClass={cs.IR},
      url={https://arxiv.org/abs/2505.17925}, 
}

@misc{shen2025optimizingsequentialrecommendationmodels,
      title={Optimizing Sequential Recommendation Models with Scaling Laws and Approximate Entropy}, 
      author={Tingjia Shen and Hao Wang and Chuhan Wu and Jin Yao Chin and Wei Guo and Yong Liu and Huifeng Guo and Defu Lian and Ruiming Tang and Enhong Chen},
      year={2025},
      eprint={2412.00430},
      archivePrefix={arXiv},
      primaryClass={cs.AI},
      url={https://arxiv.org/abs/2412.00430}, 
}

@misc{zhang2026recommendersystemsteachthemselves,
      title={Can Recommender Systems Teach Themselves? A Recursive Self-Improving Framework with Fidelity Control}, 
      author={Luankang Zhang and Hao Wang and Zhongzhou Liu and Mingjia Yin and Yonghao Huang and Jiaqi Li and Wei Guo and Yong Liu and Huifeng Guo and Defu Lian and Enhong Chen},
      year={2026},
      eprint={2602.15659},
      archivePrefix={arXiv},
      primaryClass={cs.IR},
      url={https://arxiv.org/abs/2602.15659}, 
}

@article{zhourui,
	doi = {10.20944/preprints202601.1559.v1},
	url = {https://doi.org/10.20944/preprints202601.1559.v1},
	year = 2026,
	month = {January},
	publisher = {Preprints},
	author = {Rui Zhou and Qinglin Jia and Bo Chen and Peng Xu and Yijia Sun and Siyuan Lou and Chaoxin Fu and Mengyuan Fu and Guoming Shen and Zheli Zhou and Jinlong Jiao and Naifu Zhou and Shijie Guan and Yunjing Qi and Shiyao Wang and Xinchen Luo and Qigen Hu and Chaoyi Ma and Xiao Lv and Qiang Luo and Yuyang Ye and Luankang Zhang and Defu Lian and Ruiming Tang and Guorui Zhou and Han Li and Kun Gai and Hao Wang and Enhong Chen},
	title = {A Survey of User Lifelong Behavior Modeling: Perspectives on Efficiency and Effectiveness},
	journal = {Preprints}
}

@misc{wang2025generativelargerecommendationmodels,
      title={Generative Large Recommendation Models: Emerging Trends in LLMs for Recommendation}, 
      author={Hao Wang and Wei Guo and Luankang Zhang and Jin Yao Chin and Yufei Ye and Huifeng Guo and Yong Liu and Defu Lian and Ruiming Tang and Enhong Chen},
      year={2025},
      eprint={2502.13783},
      archivePrefix={arXiv},
      primaryClass={cs.IR},
      url={https://arxiv.org/abs/2502.13783}, 
}

@misc{ye2025fuxialphascalingrecommendationmodel,
      title={FuXi-$\alpha$: Scaling Recommendation Model with Feature Interaction Enhanced Transformer}, 
      author={Yufei Ye and Wei Guo and Jin Yao Chin and Hao Wang and Hong Zhu and Xi Lin and Yuyang Ye and Yong Liu and Ruiming Tang and Defu Lian and Enhong Chen},
      year={2025},
      eprint={2502.03036},
      archivePrefix={arXiv},
      primaryClass={cs.IR},
      url={https://arxiv.org/abs/2502.03036}, 
}

@misc{wang2025universalframeworkcompressingembeddings,
      title={A Universal Framework for Compressing Embeddings in CTR Prediction}, 
      author={Kefan Wang and Hao Wang and Kenan Song and Wei Guo and Kai Cheng and Zhi Li and Yong Liu and Defu Lian and Enhong Chen},
      year={2025},
      eprint={2502.15355},
      archivePrefix={arXiv},
      primaryClass={cs.IR},
      url={https://arxiv.org/abs/2502.15355}, 
}

@misc{zhang2024unifiedframeworkadaptiverepresentation,
      title={A Unified Framework for Adaptive Representation Enhancement and Inversed Learning in Cross-Domain Recommendation}, 
      author={Luankang Zhang and Hao Wang and Suojuan Zhang and Mingjia Yin and Yongqiang Han and Jiaqing Zhang and Defu Lian and Enhong Chen},
      year={2024},
      eprint={2404.00268},
      archivePrefix={arXiv},
      primaryClass={cs.IR},
      url={https://arxiv.org/abs/2404.00268}, 
}

@misc{yu2025thoughtaugmentedplanningllmpoweredinteractive,
      title={Thought-Augmented Planning for LLM-Powered Interactive Recommender Agent}, 
      author={Haocheng Yu and Yaxiong Wu and Hao Wang and Wei Guo and Yong Liu and Yawen Li and Yuyang Ye and Junping Du and Enhong Chen},
      year={2025},
      eprint={2506.23485},
      archivePrefix={arXiv},
      primaryClass={cs.CL},
      url={https://arxiv.org/abs/2506.23485}, 
}

\clearpage
\appendix

\section{Detailed Discussion and Analysis}
\label{sec:Detailed}
\subsection{Robustness to First-Stage Retrieval and Input Ordering}
\label{sec:B.1}
To evaluate the stability of our calibration mechanism, we evaluate its performance under varying initial retrieval qualities and document orderings using \textbf{Qwen3-0.6B}. We compare three distinct settings: standard \textbf{BM25} retrieval, \textbf{Random Shuffling} of BM25 results to simulate a worst-case positional bias scenario, and the high-performance \textbf{BGE-M3} reranker \cite{chen2024bge} to test the method's ceiling.

As shown in Table \ref{tab:app_retrieval}, our calibration provides consistent and significant gains regardless of the input quality. Notably, in the \textbf{Random} setting, where the base model's performance drops due to the loss of semantic ordering, our method recovers and even surpasses the standard BM25 baseline. This demonstrates that the framework effectively decouples ranking from inherent positional artifacts in the input.

\begin{table}[h]

\centering
\footnotesize 
\setlength{\tabcolsep}{3.5pt} 
\begin{tabular}{llccccc}
\toprule
\textbf{Setting} & \textbf{Method} & \textbf{DL19} & \textbf{DL20} & \textbf{DL21} & \textbf{DL22} & \textbf{DL23} \\
\midrule
\multirow{2}{*}{BM25} 
& Base & 49.16 & 34.15 & 62.78 & 48.58 & 54.57 \\
& CapCal & \textbf{54.54} & \textbf{41.26} & \textbf{66.66} & \textbf{54.57} & \textbf{57.62} \\
\midrule
\multirow{2}{*}{Random} 
& Base & 47.61 & 32.27 & 59.07 & 47.26 & 43.60 \\
& CapCal & \textbf{56.33} & \textbf{37.57} & \textbf{65.93} & \textbf{53.16} & \textbf{52.63} \\
\midrule
\multirow{2}{*}{BGE-M3} 
& Base & 52.32 & 45.23 & 72.06 & 59.07 & 59.68 \\
& CapCal & \textbf{62.05} & \textbf{47.21} & \textbf{75.32} & \textbf{65.24} & \textbf{64.23} \\
\bottomrule
\end{tabular}
\caption{Detailed results of retrieval and ordering robustness on Qwen3-0.6B.}
\label{tab:app_retrieval}
\end{table}

\subsection{Robustness to Placeholder Content}
\label{sec:B.2}
We investigate whether the semantic content and length of the placeholder $\varnothing$ affect the estimation of positional priors. Our default \textbf{Fixed String} is the constant string ``This is a placeholder'', whose length is not matched to the original documents. We further compare empty, length-controlled, random, and passage-copy variants to assess the sensitivity of the prior estimate.

The results in Table \ref{tab:app_placeholder} show that our calibration yields improvements over the uncalibrated baseline across all variations. Overall, the results confirm that the captured prior is primarily structural rather than dependent on a specific placeholder token. We also observe that the fixed string performs particularly well on DL22 and DL23, which we conjecture is because a short but non-empty buffer helps preserve clearer separation between document identifiers than a single blank space.

\begin{table}[h]
\centering
\footnotesize
\setlength{\tabcolsep}{2.8pt}
\begin{tabular}{@{}lccccc@{}}
\toprule
\textbf{Placeholder Content} & \textbf{DL19} & \textbf{DL20} & \textbf{DL21} & \textbf{DL22} & \textbf{DL23} \\
\midrule
Base (No Calibration) & 49.16 & 34.15 & 62.78 & 48.58 & 54.57 \\
\midrule
Fixed String & 54.54 & 41.26 & 66.66 & 54.57 & 57.62 \\
Passage[1] Content & 56.12 & 40.78 & 64.05 & 50.31 & 52.98 \\
Empty (a space) & 55.82 & 44.77 & 63.50 & 50.06 & 50.48 \\
Space $\times 20$ & 57.97 & 40.43 & 64.12 & 52.96 & 53.28 \\
Random $\times 20$ & 54.15 & 38.92 & 64.58 & 49.03 & 51.43 \\
Space $\times \text{len}[1]$ & 58.38 & 43.30 & 67.30 & 52.45 & 53.65 \\
Random $\times \text{len}[1]$ & 50.24 & 40.56 & 65.10 & 51.97 & 49.29 \\
Space $\times \text{len}[i]$ & 55.95 & 42.57 & 66.29 & 52.87 & 55.86 \\
\bottomrule
\end{tabular}
\caption{Impact of different placeholder content and length on calibration performance using Qwen3-0.6B.}
\label{tab:app_placeholder}
\end{table}

\subsection{Robustness to Identifier Formats}
\label{sec:B.3}
To determine if the positional bias is a fundamental structural tendency, we switched document identifiers from numerals to alphabetic indices (e.g., ``A'', ``B''). As detailed in Table \ref{tab:app_identifier}, our calibration mechanism continues to yield clear improvements across all benchmarks with this altered format. For instance, on DL22, calibration improves the NDCG@10 from 45.60 to 53.35. This confirms that the captured positional prior is independent of specific identifier tokens, representing a generalized structural bias in the model's instruction-following behavior.

\begin{table}[h]
\centering
\footnotesize
\setlength{\tabcolsep}{3.5pt}
\begin{tabular}{lccccc}
\toprule
\textbf{Method} & \textbf{DL19} & \textbf{DL20} & \textbf{DL21} & \textbf{DL22} & \textbf{DL23} \\
\midrule
Base & 49.87 & 33.50 & 54.05 & 45.60 & 40.96 \\
CapCal & \textbf{52.16} & \textbf{36.01} & \textbf{67.70} & \textbf{53.35} & \textbf{46.80} \\
\bottomrule
\end{tabular}
\caption{Robustness results using alphabetical identifiers (A, B, C...) on Qwen3-0.6B.}
\label{tab:app_identifier}
\end{table}

\subsection{Comparison with PSC}
\label{sec:B.4}
We compare CapCal with Permutation Self-Consistency (PSC) \cite{psc}, a strong inference-time aggregation baseline that averages rankings across 10 shuffled prompts. Table \ref{tab:app_psc} shows that CapCal consistently matches or outperforms PSC on Qwen3-0.6B while using only one additional forward pass instead of 10 full reranking calls. On Qwen3-8B, the two methods are highly competitive, but CapCal remains better on DL20, DL22, DL23, Climate-FEVER, NFCorpus, and FiQA. These results indicate that directly estimating and subtracting the structural prior is a more efficient way to remove positional bias than repeatedly marginalizing over prompt orders.

\begin{table*}[t]
\centering
\footnotesize
\setlength{\tabcolsep}{4.2pt}
\begin{tabular}{llcccccccccc}
\toprule
\textbf{Model} & \textbf{Method} & \textbf{DL19} & \textbf{DL20} & \textbf{DL21} & \textbf{DL22} & \textbf{DL23} & \textbf{COVID} & \textbf{Clim.} & \textbf{NF} & \textbf{Argu.} & \textbf{FiQA} \\
\midrule
\multirow{3}{*}{Qwen3-0.6b}
& Base   & 0.4916 & 0.3415 & 0.6278 & 0.4858 & 0.5457 & 0.6074 & 0.1622 & 0.4243 & 0.1097 & 0.1807 \\
& PSC    & 0.5382 & 0.4074 & 0.6937 & 0.5246 & 0.5264 & 0.5842 & 0.1892 & 0.4047 & 0.1910 & 0.1637 \\
& CapCal & \textbf{0.5454} & \textbf{0.4126} & 0.6666 & \textbf{0.5457} & \textbf{0.5762} & \textbf{0.6400} & \textbf{0.2086} & \textbf{0.4504} & \textbf{0.2448} & \textbf{0.2364} \\
\midrule
\multirow{3}{*}{Qwen3-8b}
& Base   & 0.6807 & 0.5405 & 0.8091 & 0.6129 & 0.6945 & 0.7525 & 0.2486 & 0.4981 & 0.2436 & 0.2877 \\
& PSC    & \textbf{0.7404} & 0.5429 & \textbf{0.8473} & 0.6244 & 0.6948 & \textbf{0.7753} & 0.2627 & 0.4980 & \textbf{0.2527} & 0.2856 \\
& CapCal & 0.7141 & \textbf{0.5613} & 0.8228 & \textbf{0.6391} & \textbf{0.6979} & 0.7683 & \textbf{0.2714} & \textbf{0.5094} & 0.2464 & \textbf{0.2932} \\
\bottomrule
\end{tabular}
\caption{Comparison with Permutation Self-Consistency (PSC, $k=10$). All metrics are reported in NDCG@10.}
\label{tab:app_psc}
\end{table*}

\subsection{Limitations of Training-based De-biasing}
\label{sec:B.5}
We further investigate whether supervised fine-tuning (SFT) using permutation-augmented data can substitute for inference-time calibration. We compare \textit{Zephyr} \cite{tunstall2023zephyrdirectdistillationlm} and its augmented version \textit{RankZephyr} \cite{pradeep2023rankzephyr} against both standard and rerank-trained \textit{Qwen3-0.6B} models.

The results in Table \ref{tab:app_training} demonstrate that training-based de-biasing is not a universal solution:
\begin{itemize}
    \item \textbf{Persistence of Bias}: While rerank-training improves absolute scores for Qwen3-0.6B, the fine-tuned base model still exhibits substantial positional priors that our calibration can further rectify (e.g., DL19 improves from 61.15 to 65.53). This indicates that data augmentation alone cannot fully erase ``hard-coded'' biases in compact models.
    \item \textbf{Limits of Augmentation}: Even when training on large-scale augmented datasets, the residual bias in smaller architectures remains problematic. Unlike RankZephyr which shows saturation at a larger scale, compact models consistently benefit from explicit inference-time calibration as a necessary complement to standard training-based interventions.
\end{itemize}

\begin{table}[h]
\centering
\footnotesize
\setlength{\tabcolsep}{2.3pt}
\begin{tabular}{llccccc}
\toprule
\textbf{Model} & \textbf{Method} & \textbf{DL19} & \textbf{DL20} & \textbf{DL21} & \textbf{DL22} & \textbf{DL23} \\
\midrule
\multirow{2}{*}{Zephyr} 
& Base & 48.50 & 36.91 & 58.20 & 51.44 & 46.72 \\
& CapCal & \textbf{51.85} & 36.61 & \textbf{62.83} & \textbf{53.81} & \textbf{47.90} \\
\midrule
\multirow{2}{*}{RankZephyr} 
& Base & \textbf{76.67} & 60.73 & \textbf{89.20} & \textbf{75.11} & \textbf{76.78} \\
& CapCal & 76.25 & \textbf{60.96} & 89.15 & 74.58 & 75.55 \\
\midrule
\multirow{2}{*}{Qwen3-0.6B} 
& Base & 49.16 & 34.15 & 62.78 & 48.58 & 54.57 \\
& CapCal & \textbf{54.54} & \textbf{41.26} & \textbf{66.66} & \textbf{54.57} & \textbf{57.62} \\
\midrule
\multirow{2}{*}{\shortstack[l]{Qwen3-0.6B\\(rerank trained)}} 
& Base & 61.15 & 43.26 & 78.96 & 65.68 & 66.80 \\
& CapCal & \textbf{65.53} & \textbf{47.58} & \textbf{80.48} & \textbf{68.15} & \textbf{69.21} \\
\bottomrule
\end{tabular}
\caption{Comparative analysis between standard models and those fine-tuned with permutation-based data augmentation.}
\label{tab:app_training}
\end{table}

\newpage
\section{Ranking Prompt}
\label{sec:prompt}
\begin{figure}[htbp]
    \centering
    \noindent\fbox{%
        \begin{minipage}{\dimexpr\linewidth-2\fboxsep-2\fboxrule}
            \ttfamily 
            \small    
            \setlength{\parskip}{0.5em} 

            \textless|system|\textgreater\\
            You are RankLLM, an intelligent assistant that can rank passages based on their relevancy to the query.\\
            \textless|user|\textgreater\\
            I will provide you with \{num\} passages, each indicated by a numerical identifier []. Rank the passages based on their relevance to the search query: \{query\}.

            [1] \{passage 1\}\\
            {[2]} \{passage 2\}\\
            ...\\
            {[\{num\}]} \{passage \{num\}\}

            Search Query: \{query\}.

            Rank the \{num\} passages above based on their relevance to the search query. All the passages should be included and listed using identifiers, in descending order of relevance. The output format should be [] > [], e.g., [4] > [2]. Only respond with the ranking results, do not say any word or explain.\\
            \textless|assistant|\textgreater

            \vspace{0.5em} 

    \begin{flushleft}
        \textit{Model Generation: [9] > [4] > [20] > ... > [13]}
    \end{flushleft}
        \end{minipage}%
    }

    \caption{The reranking prompt and sample generation for listwise reranking.}
    \label{fig:listwise}
\end{figure}
\section{Additional Related Work and Potential Applications}
Although CapCal is instantiated and evaluated in ad hoc retrieval, the broader issue it addresses may arise in a wider class of generation-based decision problems. Whenever a model must select, rank, or organize multiple candidates, its behavior can be influenced not only by semantic relevance, but also by structural factors such as candidate order, identifier format, prompt layout, or uncertainty in the surrounding context. From this perspective, the problem setting studied in this work connects to a broader literature on generative ranking, retrieval-augmented generation, personalization, and explicit intervention against undesirable model biases.

\subsection{Generative Ranking and Recommendation}

A natural neighboring direction is recommendation \cite{zhourui,wang2025generativelargerecommendationmodels}, where recent work increasingly formulates retrieval, ranking, and user modeling in generative terms. Prior studies have explored unifying retrieval and ranking within a single generative recommendation framework, as well as improving personalized reranking efficiency under autoregressive generation \cite{zhang2025killingbirdsstoneunifying,cheng2026efficientpersonalizedrerankingsemiautoregressive}. More broadly, generation-based paradigms are now being investigated in foundation recommender models, reasoning-aware recommendation, long-term time-aware sequential recommendation, cross-domain user retrieval integration, recursive self-improvement, and feature-generation views of CTR modeling \cite{zhang2026thinkinghurtsdiagnosingrectifying,ye2026fuxilinearunleashingpowerlinear,ye2025fuxialphascalingrecommendationmodel,shen2024exploringuserretrievalintegration,shen2025optimizingsequentialrecommendationmodels,zhang2026recommendersystemsteachthemselves,yin2025featureinteractionfeaturegeneration,zhang2024unifiedframeworkadaptiverepresentation}. Related efforts on dynamic fusion and expert decorrelation in CTR prediction further reflect the broader challenge of producing reliable decisions over large candidate spaces, even when the task formulation differs from listwise passage reranking \cite{Wang_2025,wang2025universalframeworkcompressingembeddings,wang2025enhancingctrpredictiondecorrelated}. While these works are not direct applications of CapCal, they highlight a broader family of scenarios in which ordered or selective generation over multiple candidates is central.

\subsection{Retrieval-Augmented and Evidence-Grounded Generation}

A second relevant line of work concerns retrieval-augmented and evidence-grounded generation. Recent studies have extended retrieval-augmented generation to long-text writing, open-domain interleaved generation, continual adaptation, event extraction, and cross-lingual prompting or in-context generation \cite{gu2025rapidefficientretrievalaugmentedlong,zhang2025ragigbenchinnovativeevaluationragbased,huang2025selfaugmitigatingcatastrophicforgetting,liang-etal-2025-adaptive,nie-etal-2023-cross,li-etal-2023-from}. Although these tasks are not listwise reranking in the strict sense, they often require the model to identify, prioritize, filter, or organize multiple retrieved candidates before or during generation. In such settings, the final output quality may depend not only on what evidence is retrieved, but also on how candidate evidence is exposed to the model. This makes them conceptually relevant when thinking about structural priors that can distort generation-time decisions.

\subsection{Personalization, Memory, and User-Centric Agents}

The same concern may become more pronounced in personalized and agentic systems, where candidate selection is repeatedly performed under evolving user context. Recent work has explored decoding-time personalization through collaborative steering, efficient personalized generation under edge--cloud collaboration, and broader user-centric agent paradigms in which personalized decision modules mediate between private user context and external information services \cite{lv2026costeercollaborativedecodingtimepersonalization,lv2026specsteersynergizinglocalcontext,zhang2026paradigmusercentricagentplatformcentric,yu2025thoughtaugmentedplanningllmpoweredinteractive}. At the same time, memory has emerged as an increasingly important abstraction for long-horizon LLM systems \cite{wu-etal-2025-from}. In these settings, ranking or evidence prioritization is often not a one-shot operation, but part of an ongoing interaction loop. This raises the possibility that structural artifacts, if left uncorrected, may accumulate over time and affect the consistency of downstream behavior.

\subsection{Structural Biases and Explicit Intervention}

More broadly, CapCal is aligned with a general line of research that treats undesirable model behavior as something to be explicitly identified and corrected, rather than passively hoped away. Prior work has examined bias reduction in contextualized representations, as well as data-utility-aware alignment and curriculum strategies for LLM optimization \cite{liang-etal-2020-monolingual,zhi2026spardselfpacedcurriculumrl}. These works address different phenomena from the position bias studied here, but they share a useful methodological intuition: model outputs can be systematically skewed by factors that are not part of the target semantic signal, and explicit intervention may therefore be preferable to relying on implicit adaptation alone.

Taken together, these directions suggest that content-agnostic calibration may be viewed not merely as a retrieval-specific technique, but as one concrete instance of a broader research theme: understanding and correcting structural distortions in generation-based decision processes. Exploring such extensions is beyond the scope of the present paper, but we believe it is a promising direction for future work.

\label{sec:appendix}

\end{document}